\documentclass[twocolumn]{svproc}

\usepackage{url}
\usepackage{comment}

\usepackage[super,numbers]{natbib}
\usepackage{url} % Add to preamble
\usepackage{amsmath} % Required for equations
\usepackage{amssymb}
\usepackage[a4paper, 
            left=1.5cm, 
            right=1.5cm, 
            top=1.5cm, 
            bottom=1.5cm, % Adjust this for bottom margin
            footskip=1cm,]{geometry}
\usepackage{graphicx}
\usepackage{subcaption}
\usepackage{wrapfig}
\usepackage{tcolorbox}     % For advanced boxing
\tcbuselibrary{breakable}  % For page breaks inboxes
\usepackage[font=small]{caption}
\usepackage{tabularx}
\usepackage{amsfonts} 
\usepackage{enumitem}
\usepackage{booktabs}
\usepackage{array}
\begin{document}
\mainmatter             
\title{SAGE-GAN: Towards Realistic and Robust Segmentation of Spatially Ordered Nanoparticles via Attention-Guided GANs}
\titlerunning{Nanoparticle Image Segmentation}  
\author{Anindya Pal\inst{1}\thanks{These authors contributed equally.} \and Varun Ajith\inst{1}$^\star$,
Saumik Bhattacharya \inst{2} \and Sayantari Ghosh \inst{1}}
\authorrunning{Pal et al.} % abbreviated author list (for running head)

\institute{National Institute of Technology Durgapur, Department of Physics\\
\email{sghosh.phy@nitdgp.ac},
\and
Indian Institute of Technology Kharagpur}
%
%
%
%
% Put abstract in the optional argument of \twocolumn
\twocolumn[
  \maketitle
    \begin{abstract}
Precise analysis of nanoparticles for characterization in electron microscopy images is essential for advancing nanomaterial development. Yet it remains challenging due to the time-consuming nature of manual methods and the shortcomings of traditional automated segmentation techniques, especially when dealing with complex shapes and imaging artifacts. While conventional methods yield promising results, they depend on a large volume of labeled training data, which is both difficult to acquire and highly time-consuming to generate. In order to overcome these challenges, we have developed a two-step solution: Firstly, our system learns to segment the key features of nanoparticles from a dataset of real images using a self-attention driven U-Net architecture that focuses on important physical and morphological details while ignoring background features and noise.  Secondly, this trained Attention U-Net is embedded in a cycle-consistent generative adversarial network (CycleGAN) framework, inspired by the cGAN-Seg model introduced by Abzargar et al. This integration allows for the creation of highly realistic synthetic electron microscopy image-mask pairs that naturally reflect the structural patterns learned by the Attention U-Net. Consequently, the model can accurately detect features in a diverse array of real-world nanoparticle images and autonomously augment the training dataset without requiring human input. Cycle consistency enforces a direct correspondence between synthetic images and ground-truth masks, ensuring realistic features, which is crucial for accurate segmentation training. Visualizations of attention overlays on the validation set, comparison of real images with their corresponding attention maps, show that the model achieves precise segmentation by concentrating on morphologically relevant regions. Unlike simpler simulation methods or standard image generators, our approach ensures that the synthetic images maintain accurate feature and shapes. Experimental validation on real electron microscopy images demonstrated excellent segmentation accuracy and robustness, particularly under data-limited conditions and complicated morphologies.
\end{abstract}
]
\section{Introduction}
\begin{figure*}[!ht]
%\vspace{-1.7\baselineskip} % Pull figure up into paragraph
\centering
\includegraphics[width=1\linewidth]{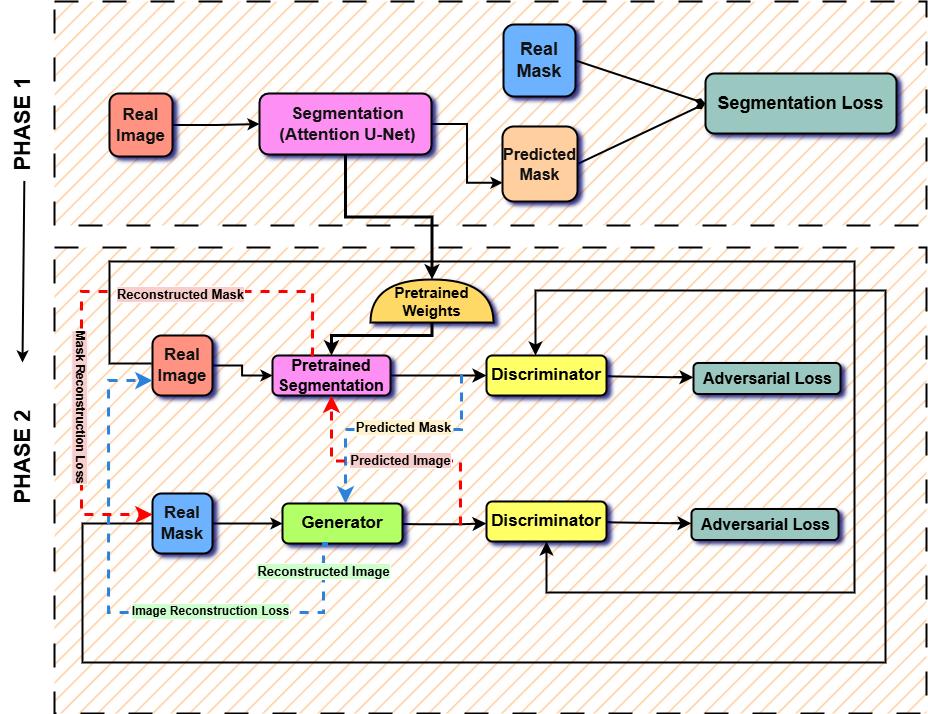}
\caption{Schematic of the proposed two phase pipeline. Phase one depicts the pretraining of the Attention U-Net architecture and Phase two represents the training of the CycleGAN framework with the pretrained U-Net embedded within the architecture. The broken arrows indicate the cycle consistency between the generator and the segmentation network.}
\label{Fig:model}
\end{figure*}

Nanoparticles, typically defined as particles with at least one dimension between 1 and 100 nanometers exhibit distinctive physical, chemical, and biological properties that set them apart from their bulk counterparts. These unique characteristics stem from their high surface-to-volume ratio, quantum effects, and size-dependent optical behaviors. Such nanoscale properties, including enhanced reactivity and tunable optical responses, form the basis for their diverse applications in fields like medicine, energy storage, sensing, drug delivery, and materials science \cite{dreaden2012golden,lohse2012applications,Vance2015,sun2007bright,zhang2014near}. As research continues, nanoparticles are expected to lead major advances in fields like clean energy and personalized healthcare. Their growing use in different industries shows how important they have become for technological progress and daily life.
\vspace{4pt}\\
Accurate characterization is essential for understanding and leveraging the unique properties of nanoparticles, such as their size, shape, surface chemistry, crystallinity, and degree of aggregation. However, realizing their full potential requires precise structural characterizations. These characteristics directly influence the behavior and performance of nanomaterials in various applications, ranging from cosmetics, textiles, food and beverages to medicine and healthcare. Without precise characterization, it becomes challenging to control or predict these behaviors, ultimately limiting the development of high-performance nanomaterials tailored for specific industrial, biomedical, or energy applications. Among various techniques, electron microscopy (EM) methods such as Scanning Electron Microscopy (SEM) \cite{suga2014recent} and Transmission Electron Microscopy (TEM) \cite{anjum2016characterization} remain standard due to their ability to capture fine structural details, including particle size, shape, and spatial distribution. However, manual analysis of these images is labor intensive, subjective, and error-prone, particularly for complex nanostructures with overlapping particles or heterogeneous morphologies. The growing complexity of nanomaterials and the increasing resolution of microscopy techniques further amplify the need for robust computational tools that can reliably extract quantitative data from images while minimizing human intervention. These challenges create a need for automated image analysis techniques that can deliver fast, objective, and reproducible results.
\vspace{4pt}\\
With the advancement of machine learning-driven image analyses through the last decade, automated image segmentation has emerged as a critical tool for quantitative nanoparticle analysis. Conventional approaches, including thresholding \cite{groom2018automatic}, edge detection \cite{meng2018automatic} and watershed algorithms \cite{li2018automated}, struggle with challenges such as low contrast, imaging artifacts, and agglomerated particles. Recent advances in deep learning, particularly U-Net-based architectures \cite{ronneberger2015unetconvolutionalnetworksbiomedical}, have improved segmentation accuracy by learning hierarchical features directly from the data. For example, Mask R-CNN \cite{monchot2021deep} and lightweight networks such as NSNet \cite{sun2022deep} demonstrate promising results in separating nanoparticles. However, these models require large, diverse, and meticulously annotated datasets to generalize effectively. This remians as a major bottleneck in EM imaging, where acquiring and labeling high-resolution images is time-consuming, costly, and domain specific.
\vspace{4pt}\\
To address these challenges, the use of synthetic data generation has recently gained significant attention. Liang et al. \cite{liang2024segmentation} augmented real TEM datasets with synthetic image-mask pairs, achieving optimal segmentation with only 15\% authentic data. Their core idea was to augment the real dataset with the incorporation of software-generated image mask pairs. However, their model required critical human supervision and guidance throughout the creation of synthetic images. The model also failed to incorporate crucial physical details present in the actual dataset. Shah et al. \cite{shah2023automated} applied DeepLIFT analysis \cite{shrikumar2017learning} to study 2D materials like graphene, highlighting several distinct challenges. These included inconsistent sizes and shapes, varying contrast ratios, complex background features, and a shallow depth of field. The authors noted that the model struggled to replicate the natural variations found in real microscopy images. Mill et al. \cite{mill2021synthetic} proposed a segmentation workflow for complex nanoparticle structures in high-resolution microscopy images using a semi-automated synthetic data pipeline based on realistic rendering using software such as Blender \cite{Blender}. Their approach was able to generate an unlimited number of synthetic data with the corresponding ground truth mask, allowing training of convolutional neural networks(CNNs) \cite{oshea2015introductionconvolutionalneuralnetworks}. However, these methods were based on manual template design or rigid physical models, which lacked the diversity and realism needed to capture the full complexity of experimental electron microscopy images. This trade-off between scalability and realism restricted their utility for analyzing heterogeneous or irregular nanostructures. Furthermore, vanilla generative adversarial networks (GANs) \cite{goodfellow2014generativeadversarialnetworks}, while capable of producing realistic synthetic images, often failed to preserve structural fidelity to ground-truth masks, limiting their value for training segmentation models.
\vspace{4pt}\\
In this paper, we have introduced a two-phase architecture (Fig-\ref{Fig:model}) that integrates an attention-guided \cite{vaswani2017attention} U-Net with a cycle-consistent generative adversarial network (CycleGAN) \cite{zhu2020unpairedimagetoimagetranslationusing} inspired by the cGAN-Seg model \cite{zargari2024enhanced} which was introduced to address data scarcity without sacrificing morphological accuracy.
In \textit{Phase 1}, the Attention U-Net \cite{oktay2018attentionunetlearninglook} is first trained on the available real SEM images and their corresponding manually annotated masks. This stage focuses on learning segmentation by leveraging spatial attention gates to prioritize regions of interest (e.g., nanoparticle boundaries and spatial features). The trained model captures domain-specific structural features, such as size, shape, and spatial distribution, while mitigating noise and artifacts through its attention mechanism. Next, in \textit{Phase 2}, the trained Attention U-Net \cite{oktay2018attentionunetlearninglook} is integrated into a CycleGAN \cite{zhu2020unpairedimagetoimagetranslationusing} framework to enforce cycle consistency. By integrating the trained U-Net with the CycleGAN \cite{zhu2020unpairedimagetoimagetranslationusing} framework, the generator learns to produce synthetic images that align with the structural priors encoded in the Attention U-Net \cite{oktay2018attentionunetlearninglook}. In addition, adversarial training ensures that synthetic images mimic the texture, noise, and contrast variations of real SEM data.
Unlike previous synthetic data approaches, our method leverages cycle consistency to maintain structural correspondence between synthetic images and ground-truth masks, ensuring that generated data aligns with physical reality. The attention mechanism further enhances segmentation accuracy by prioritizing morphologically relevant regions, reducing errors from imaging artifacts. This approach not only mitigates data scarcity but also provides a pathway for adaptive model retraining as new systems emerge.
\vspace{4pt}\\
The rest of the paper is structured as follows: Section \ref{sec: meth} explains our approach, including the architectural details and methodology to follow. Section \ref{sec: data} provides information on the dataset used to train our model. Section \ref{sec: eval_met} describes the choices for the evaluation metrics. Section \ref{sec: results} shows quantitative and qualitative results, followed by code availability (Section \ref{sec: code}), training configuration (Section \ref{sec: trainconfig}) and conclusion in Section \ref{sec: conclusion}. Our work introduces a scalable pipeline for nanoparticle analysis, offering potential to accelerate nanomaterial characterization and improve quality control processes.

\section{Methodology}
\label{sec: meth}
\subsection{Attention U-Net}
\begin{figure}[htbp]
%\vspace{-1.7\baselineskip} % Pull figure up into paragraph
\centering
\includegraphics[width=0.85\linewidth]{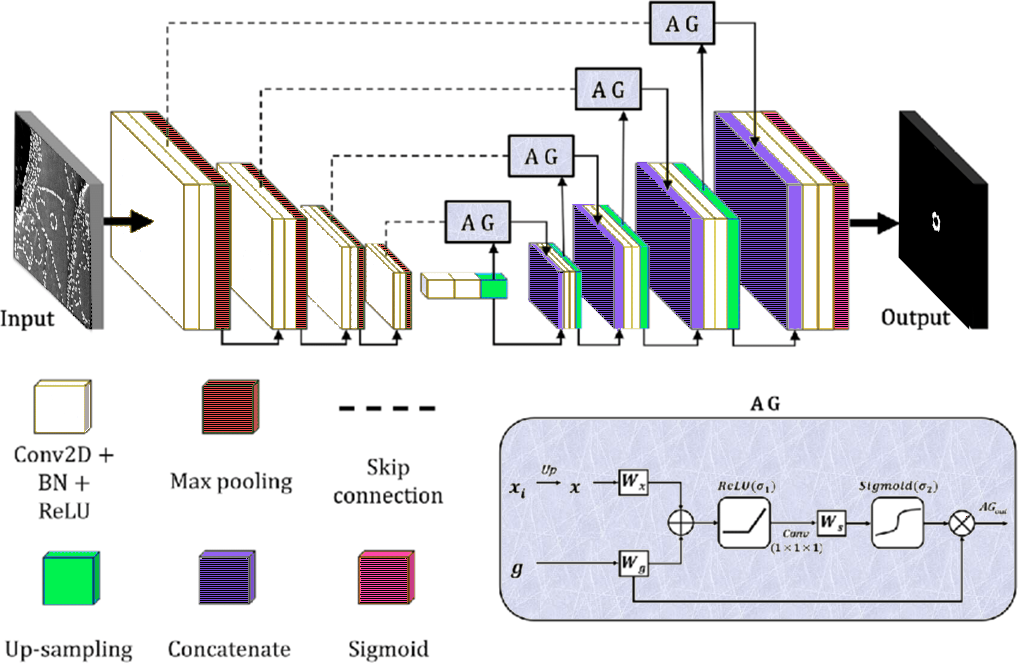}
\caption{Attention U-Net}
\label{Fig: attn_unet}
\end{figure}

The \textbf{Attention U-Net} \cite{oktay2018attentionunetlearninglook} is an advanced variant of the traditional U-Net architecture, designed to improve image segmentation, particularly in biomedical imaging, by incorporating attention gates (AGs). Although the standard U-Net uses skip connections that concatenate encoder and decoder features, it cannot distinguish between spatial regions of interest and irrelevant background, leading to noisy decoding. Attention U-Net \cite{oktay2018attentionunetlearninglook} addresses this limitation by learning to focus on salient regions via AGs, suppressing irrelevant features and improving segmentation accuracy in complex, low-contrast, or heterogeneous scenes.
\vspace{2pt}\\
The backbone of Attention U-Net \cite{oktay2018attentionunetlearninglook} remains the same encoder-decoder structure of U-Net \cite{ronneberger2015unetconvolutionalnetworksbiomedical}: convolution and pooling layers encode hierarchical features, and upsampling layers decode them. Attention gates are applied before skip connections to generate \textit{attention coefficients} \cite{vaswani2017attention} that weight encoder features conditionally according to the decoder context. The schematic of the Attention U-Net architecture is shown in figure-\ref{Fig: attn_unet}.
\subsubsection*{Mathematical Formulation of Attention Gates\vspace{2pt}\\}
Let the encoder feature map be:
\begin{equation}
x \in \mathbb{R}^{F_l \times H \times W}
\end{equation}
and the gating signal from the decoder be:
\begin{equation}
g \in \mathbb{R}^{F_g \times H' \times W'}
\end{equation}
The goal is to generate an attention map:
\begin{equation}
\alpha \in [0, 1]^{1 \times H \times W}
\end{equation}
that selectively weights the encoder features spatially.
%\vspace{2pt}\\
First, we perform Linear Transformations, where inputs are linearly projected to a common feature space using $1 \times 1$ convolutions:
    \begin{equation}
    \theta_x = W_x x, \quad \phi_g = W_g g
    \end{equation}
    where: 
    \[
    W_x \in \mathbb{R}^{F_{\text{int}} \times F_l}, \quad
    W_g \in \mathbb{R}^{F_{\text{int}} \times F_g}
    \]

    Assume $\theta_x, \phi_g \in \mathbb{R}^{F_{\text{int}} \times H \times W}$ after upsampling $\phi_g$ as necessary to match the spatial dimensions.\\
Next, element-wise addition followed by a non-linear activation:
    \begin{equation}
    f = \text{ReLU}(\theta_x + \phi_g)
    \end{equation}
After the additive Attention has been incorporated, Attention coefficient computation is done, where a $1 \times 1$ convolution followed by a sigmoid activation gives the attention map:
    \begin{equation}
    \alpha = \sigma(\psi^T f + b_\psi)
    \end{equation}
    where:
    \[
    \psi^T \in \mathbb{R}^{1 \times F_{\text{int}}}, \quad \alpha \in [0, 1]^{1 \times H \times W}
    \]\\
Finally, the encoder feature map is scaled element-wise by the attention coefficients:
    \begin{equation}
    \hat{x}_i = \alpha_i \cdot x_i
    \end{equation}
    yielding the attended feature map $\hat{x}$, which is then passed to the decoder via the skip connection.
This formulation enables the model to selectively transmit spatial features conditioned on decoder context, effectively acting as a soft spatial attention mechanism. By suppressing irrelevant background and enhancing feature representations of the target structures, Attention U-Net \cite{oktay2018attentionunetlearninglook} significantly improves segmentation accuracy, especially in tasks involving complex and variable anatomical regions. 
\subsection{CycleGAN}
\begin{figure}[!ht]
%\vspace{-1.7\baselineskip} % Pull figure up into paragraph
\centering
\includegraphics[width=0.9\linewidth]{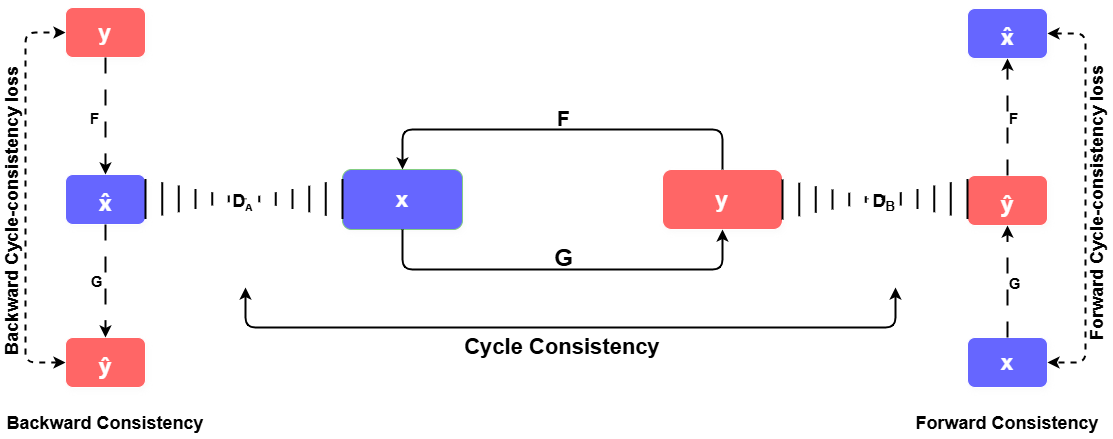}
\caption{Cycle GAN}
\label{Fig:cycleGAN}
\end{figure}
Let $\mathcal{D}_A$ and $\mathcal{D}_B$ be two image domains with no paired correspondence. CycleGAN \cite{zhu2020unpairedimagetoimagetranslationusing} is a generative framework designed for unpaired image-to-image translation by learning mappings between these domains using adversarial training and cycle consistency (shown in figure-\ref{Fig:cycleGAN}). The architecture comprises two generators:
\[
G: \mathcal{D}_A \rightarrow \mathcal{D}_B, \quad F: \mathcal{D}_B \rightarrow \mathcal{D}_A
\]
and two discriminators $D_A$ and $D_B$, each trained to distinguish between real and generated samples in their respective domains.
\subsubsection*{Adversarial Objective\vspace{2pt}\\}
The adversarial losses for domains $\mathcal{D}_A$ and $\mathcal{D}_B$ are defined as:
\begin{equation}
\begin{split}
\mathcal{L}_{\text{GAN}}(G, D_B, \mathcal{D}_A, \mathcal{D}_B) =\mathbb{E}_{y \sim p_{\mathcal{D}_B}(y)}[\log D_B(y)]\\ + \mathbb{E}_{x \sim p_{\mathcal{D}_A}(x)}[\log(1 - D_B(G(x)))]
\end{split}
\end{equation}
\begin{equation}
\begin{split}
\mathcal{L}_{\text{GAN}}(F, D_A, \mathcal{D}_B, \mathcal{D}_A) = \mathbb{E}_{x \sim p_{\mathcal{D}_A}(x)}[\log D_A(x)] \\+ \mathbb{E}_{y \sim p_{\mathcal{D}_B}(y)}[\log(1 - D_A(F(y)))]
\end{split}
\end{equation}
\subsubsection*{Cycle Consistency Loss\vspace{2pt}\\}
To enforce that the mappings are cycle-consistent, the following cycle-consistency loss is introduced:
\begin{equation}
\begin{split}
\mathcal{L}_{\text{cyc}}(G, F) = \mathbb{E}_{x \sim p_{\mathcal{D}_A}(x)} \left[\|F(G(x)) - x\|_1 \right] \\+ \mathbb{E}_{y \sim p_{\mathcal{D}_B}(y)} \left[\|G(F(y)) - y\|_1 \right]
\end{split}
\end{equation}
\subsubsection*{Full Objective\vspace{2pt}\\}
The overall objective function for the CycleGAN model is:
\begin{equation}
\begin{split}
\mathcal{L}_{\text{total}} = \mathcal{L}_{\text{GAN}}(G, D_B, \mathcal{D}_A, \mathcal{D}_B)+\lambda \mathcal{L}_{\text{cyc}}(G, F)\\+\mathcal{L}_{\text{GAN}}(F, D_A, \mathcal{D}_B, \mathcal{D}_A)
\end{split}
\end{equation}
where $\lambda$ is a regularization parameter controlling the relative importance of the cycle-consistency loss.
\subsubsection*{Application in Synthetic Data Generation\vspace{2pt}\\}
In our application, domain $\mathcal{D}_A$ consists of binary mask images representing annotated nanoparticles, and domain $\mathcal{D}_B$ contains real or synthetic nanoparticle microscopy images. The generator $F$ translates an image $I_B \in \mathcal{D}_B$ to a synthetic mask $I_A = F(I_B)$, while $G$ attempts to reconstruct the original image $\hat{I}_B = G(I_A)$. This bidirectional consistency  ensures preservation of both local and global semantic content across domain transformations.
\subsubsection*{Segmentation Augmentation Strategy\vspace{2pt}\\}
To enhance segmentation performance under limited data conditions, we adopt an online augmentation strategy during the main training phase of the CycleGAN \cite{zhu2020unpairedimagetoimagetranslationusing} architecture, which embeds the pretrained Attention U-Net. Instead of statically augmenting the training dataset, synthetic images are generated dynamically during training using the generator \( G \), such that:
\begin{equation}
\begin{split}
\mathcal{D}_{\text{synthetic}}(t) = \{(G(I_A), I_A) \mid I_A \in \mathcal{D}_A \\ \text{ at iteration } t\}
\end{split}
\end{equation}
At each training iteration \( t \), the model is exposed to both real samples from \( \mathcal{D}_{\text{real}} \) and synthetic samples from \( \mathcal{D}_{\text{synthetic}}(t) \). This results in an effective training set:
\begin{equation}
\mathcal{D}_{\text{effective}}(t) = \mathcal{D}_{\text{real}} \cup \mathcal{D}_{\text{synthetic}}(t),
\end{equation}
This dual exposure to real and synthetic data during training effectively simulates a larger and more diverse dataset, improving the model's generalization capability and robustness to variations in structure, contrast, and noise. Unlike static dataset augmentation, this method provides continual, varied input that mitigates overfitting and enhances downstream segmentation performance.

\subsection{Style-Based U-Net Generator}
The generator $\mathcal{G}$ is based on a modified 2D U-Net \cite{ronneberger2015unetconvolutionalnetworksbiomedical} architecture, designed to synthesize high-fidelity S1 nanoparticle images. The U-Net framework consists of an encoder–decoder structure $\mathcal{G}: z \rightarrow \hat{x}$, where the encoder captures semantic content, and the decoder reconstructs spatially coherent outputs.
\vspace{2mm}\\
To enrich generative diversity while preserving structural fidelity, we integrate a style decoding branch into the decoder, inspired by the cGAN-Seg \cite{zargari2024enhanced} model. The style generation pipeline introduces a mapping network $M: \mathcal{Z} \rightarrow \mathcal{W}$, where $z \in \mathcal{Z}$ is a latent noise vector, and $w = M(z) \in \mathcal{W}$ is the corresponding style vector. The decoder layers are modulated by the style vector $w$ using Adaptive Instance Normalization (AdaIN). At each layer $\ell$, feature maps $F_\ell$ are normalized and transformed as:
\[\
\text{AdaIN}(F_\ell, w) = \sigma(w_\ell) \cdot \frac{F_\ell - \mu(F_\ell)}{\sigma(F_\ell)} + \mu(w_\ell),
\]
where $\mu(\cdot)$ and $\sigma(\cdot)$ denote the mean and standard deviation, respectively, and $w_\ell$ is a learned affine transformation of $w$ for layer $\ell$.
\vspace{2mm}\\
Additionally, spatial noise $\epsilon_\ell \sim \mathcal{N}(0, I)$ is injected into selected layers to simulate fine-grained texture variations, thereby mimicking the heterogeneity inherent to real S1 nanoparticle samples.
\vspace{2mm}\\
This dual-branch architecture, termed \textit{Style-U-Net} \cite{karras2019stylebasedgeneratorarchitecturegenerative}, achieves a balance between spatial accuracy and generative richness. The content is governed by the encoder's semantic features, while the style is controlled by modulation and noise perturbations, enabling synthesis of topologically consistent yet texturally diverse images with realistic variations.
\subsection{Training Workflow}
Our approach aims to enhance segmentation performance when labeled data is scarce by combining synthetic data generation with segmentation-aware training. The workflow consists of two main stages:

\begin{enumerate}
    \item Pre-training the Attention U-Net \cite{oktay2018attentionunetlearninglook} segmentation network.
    \item Integration of the pre-trained segmentation weights into CycleGAN \cite{zhu2020unpairedimagetoimagetranslationusing} training.
\end{enumerate}
\subsubsection*{Attention U-Net Segmentation Pretraining\vspace{2pt}\\}
In the first stage, the Attention U-Net \cite{oktay2018attentionunetlearninglook} model is trained on the available labeled dataset to perform semantic segmentation. Attention gates within the network allow the model to focus on the most relevant regions in the images, enhancing feature extraction by suppressing irrelevant or noisy background information. This supervised training enables the segmentation network to learn both low-level and high-level features that capture the anatomical or structural details required for accurate segmentation. The pre-trained weights obtained here serve as foundational representation and are later used to guide the synthetic data generation process.
\subsubsection*{Integration of the Pretrained Segmentation network into CycleGAN Training\vspace{2pt}\\}
The novel aspect of our framework is the integration of the pretrained Attention U-Net \cite{oktay2018attentionunetlearninglook} segmenter into the CycleGAN \cite{zhu2020unpairedimagetoimagetranslationusing} training loop. After a synthetic image is generated, it is passed through the segmentation network to obtain a predicted segmentation mask. This predicted mask is then used to evaluate the structural quality of the generated image, providing feedback to the generator. By incorporating this segmentation-aware feedback, the generator is encouraged to produce images that are not only visually realistic but also maintain meaningful structural features relevant to segmentation.
%\vspace{2pt}\\
This segmentation-guided training helps the model generate synthetic images that contribute effectively to training downstream segmentation networks, improving generalization and robustness despite limited original labeled data. The qualitative fidelity of the images generated by the CycleGAN \cite{zhu2020unpairedimagetoimagetranslationusing} component is illustrated in Fig-S1\footnote{For figure-S1 refer to supplementary material.}.

\subsection{Loss Functions}
\subsubsection*{Generation Losses\vspace{2pt}\\}
The model integrates two complementary loss functions, VGG-based perceptual loss \cite{johnson2016perceptuallossesrealtimestyle} and pixel-wise $\mathcal{L}_1$ loss to improve the realism of generated images. The perceptual loss utilizes a pre-trained VGG network to compare high-level feature representations between synthetic and real images, promoting textural consistency and perceptual fidelity. In contrast, the $\mathcal{L}_1$ loss operates at the pixel level, minimizing absolute differences between generated and ground-truth images, thereby preserving critical structural attributes such as nanoparticle size and orientation.
\subsubsection*{Discriminator Loss\vspace{2pt}\\}
The Mean Squared Error (MSE) loss is used as the discriminator loss for both the mask discriminator and the image discriminator, providing stable and consistent feedback to the generator. MSE penalizes large errors more harshly than L1, promoting smoother convergence and deterring erratic updates during training. For the image discriminator, this loss encourages pixel-level realism in generated $\mathrm{S1}$ nanoparticles, while the mask discriminator uses MSE to encourage precise segmentation boundaries in synthetic masks. This choice aligns with your goal of generating high-fidelity, structurally accurate outputs, as MSE's geometric alignment emphasis complements the generator's perceptual (VGG) and pixel-wise (L1) losses.
\subsubsection*{Segmentation Loss\vspace{2pt}\\}
Considering the size and coverage of the $S1$ nanoparticles in the images in the dataset, we employed a combination of \textit{Binary Cross-Entropy Loss} and \textit{Focal Tversky Loss} \cite{abraham2018novelfocaltverskyloss}, ensuring a balanced approach to predict accurate pixel level classification and tackle the inherent class imbalance present in the dataset. The class-balanced focal cross-entropy loss is defined as:
\begin{equation}
\mathcal{L}_{\text{CE}} = -\log(p_t)
\label{eq:fce}
\end{equation}
The boundary-aware Focal Tversky loss is computed as:
\begin{equation}
\mathcal{L}_{\text{TV}} = (1 - \text{TI})^{\gamma}
\label{eq:ft}
\end{equation}
where the Tversky Index (TI) measures segmentation overlap as $\text{TI} = \frac{TP}{TP + \alpha FP + \beta FN}$, with $TP$, $FP$, $FN$ denoting true positives, false positives, false negatives,  $\alpha$ \& $\beta$ controlling the false positive/negative trade-off and $\gamma$ controlling the down-weighting of easy samples.
\vspace{2mm}\\
The total segmentation loss is defined as:
\begin{equation}
\mathcal{L}_{\text{Total}} = \lambda_1 \mathcal{L}_{\text{CE}} + \lambda_2 \mathcal{L}_{\text{FTV}}
\label{eq:total_loss}
\end{equation}
where $\lambda_1$ and $\lambda_2$ are relative weighting factors influencing the contribution of each of these respective components towards the entire loss function. In order to achieve optimal results we assigned equal weightage with $\lambda_1 = \lambda_2 = 1$.
\section{Dataset and Pre-processing}
\label{sec: data}
\subsection{Dataset Description}
\begin{figure}[!ht]
    \centering
    
    % Single pair of images
    \begin{minipage}[t]{0.49\linewidth}
        \centering
        \includegraphics[width=0.95\linewidth]{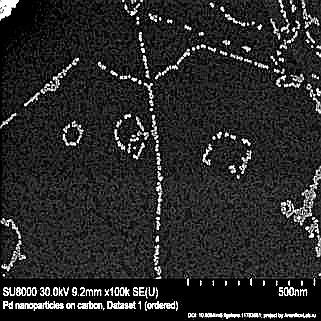}
        \subcaption{Image}
    \end{minipage}
    \hfill
    \begin{minipage}[t]{0.49\linewidth}
        \centering
        \includegraphics[width=0.95\linewidth]{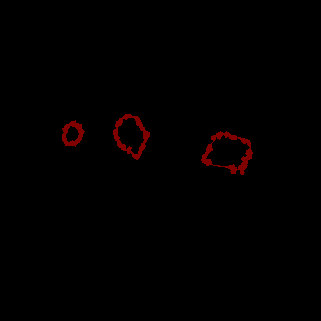}
        \subcaption{Mask}
    \end{minipage}
    
    \caption{Sample image and its corresponding ground truth label from the original S1 dataset}
    \label{fig:datasample}
\end{figure}

The dataset used in this study comprises a total of 540 labeled images representing S1 nanoparticle structures  \cite{boiko2020electron} which have unique spatial order and specific geometrical patterns. It includes 140 authentic Scanning Electron Microscopy (SEM) images and 400 synthetic images generated using K-3D  \cite{K3D}, a 3D modeling software. The dataset is publically available and can be obtained from the following link -  (\url{https://doi.org/10.1371/journal.pone.0311228.s001}) . For the purpose of ensuring high fidelity in model training and evaluation, only the 140 authentic images— hereafter referred to as AD were utilized in all training and validation procedures. A sample pair of real image and corresponding mask is shown on Fig-\ref{fig:datasample}. These images were sourced from the publicly available Boiko dataset   \cite{boiko2020electron} labeled by Liang et al.  \cite{liang2024segmentation}.
\begin{comment}
\begin{figure}[!ht]
\centering
\begin{subfigure}{0.49\textwidth}
\centering\includegraphics[width=0.5\textwidth]{data_sample_real_image.png}
\caption{Image}
\end{subfigure}
\centering
\begin{subfigure}{0.49\textwidth}
\centering\includegraphics[width=0.5\textwidth]{data_sample_real_mask.png}
\caption{Mask}
\end{subfigure}
\caption{Sample image and its corresponding ground truth label from the original S1 dataset}
\label{fig: datasample}
\end{figure}
\end{comment}
%%%%%
%%%%%
\subsection{Training and Validation Split}

The 140 authentic images were divided into training and validation sets using an 80:20 ratio. This resulted in 112 training samples and 28 validation samples. The partitioning strategy was chosen to provide sufficient data for learning while preserving a robust validation set to monitor generalization performance and avoid overfitting.

\subsection{Dataset Preprocessing}
To facilitate consistency in input data for the neural network, the SEM images were pre-processed using methods like scaling and normalization. Resizing scales the images to a constant size, and normalization rescales pixel values to match the needs of the network, thus maximizing computational effectiveness. These normalization processes make the model more resilient and capable in dealing with variations in nanoparticle morphology usually found in SEM images. Augmentation techniques like horizontal and vertical flipping, CLAHE, random crop etc. were employed to ensure that the model stays robust to variations and identifies the important patterns in data distributions that might vary from the training dataset in real world

\subsection{Exclusion of Synthetic Data}

Synthetic images were intentionally excluded from direct use in the training process, in line with contemporary augmentation strategies that emphasize the quality and realism of data over quantity. While synthetic data can enhance datasets, overdependence on manually created images risks introducing domain shifts, leading models to overfit to artificial features rather than capturing the genuine variability found in real microscopy data. Instead, our approach leverages synthetic data indirectly through the generative component of the proposed model, which learns to represent the distribution within the authentic data (AD) more subtly and effectively during training.

\section{Evaluation Metrics}
\label{sec: eval_met}
\subsection*{Dice Similarity Coefficient (Dice Score)}
The Dice score quantifies the spatial agreement between the predicted segmentation mask $P$ and the ground truth $G$, particularly suitable for nanoparticle analysis where the preservation of particle morphology and coverage is crucial:

\begin{equation}
\begin{split}
    \text{Dice}(P,G) = \frac{2|P \cap G|}{|P| + |G|} = \frac{2\sum_{i=1}^{N} p_i g_i}{\sum_{i=1}^{N} p_i + \sum_{i=1}^{N} g_i} \\= \frac{2TP}{2TP + FP + FN}
\end{split}
\end{equation}

where:
\begin{itemize}[label=$\bullet$]
    \item $p_i \in \{0,1\}$: Binary prediction at pixel $i$
    \vspace{1mm}
    \item $g_i \in \{0,1\}$: Ground truth value at pixel $i$
    \vspace{1mm}
    \item $N$: Total number of pixels in the image
    \vspace{1mm}
    \item $TP = \sum_{i=1}^{N} p_i g_i$ (True Positives)
    \vspace{1mm}
    \item $FP = \sum_{i=1}^{N} p_i (1-g_i)$ (False Positives)
    \vspace{1mm}
    \item $FN = \sum_{i=1}^{N} (1-p_i) g_i$ (False Negatives)
\end{itemize}
\vspace{2mm}
%%%%%%%
\subsection*{F-Score (F1 Measure)}
The F1-score provides a balanced measure considering both precision and recall, with adjustable emphasis via the $\beta$ parameter:

\begin{equation}
    \text{F1-score} = \left(1 + \beta^2\right) \frac{\text{Precision} \cdot \text{Recall}}{\beta^2 \cdot \text{Precision} + \text{Recall}}
\end{equation}
we use $\beta=1$ (F1-score) to equally weight precision and recall
\begin{equation}
    \text{F1} = 2 \cdot \frac{\text{Precision} \cdot \text{Recall}}{\text{Precision} + \text{Recall}} = \frac{TP}{TP + \frac{1}{2}(FP + FN)}
\end{equation}

where:
\begin{itemize}[label=$\bullet$]
    \item $\text{Precision} = \frac{TP}{TP + FP}$
    \item $\text{Recall} = \frac{TP}{TP + FN}$
\end{itemize}
The Dice score and F1-score range from 0 to 1, where 1 indicates perfect overlap between predicted and ground truth masks, and 0 indicates no overlap. These metrics evaluate the accuracy of segmentation models, with higher values reflecting better performance in identifying true positive regions.

\subsection*{Theoretical Relationship}
For binary segmentation, the Dice score and F1-score are numerically equivalent:

\begin{equation}
    \text{Dice}(P,G) \equiv \text{F1}(P,G)
\end{equation}

However, their implementations differ in:
\begin{itemize}
    \item \textit{Smoothing}: Dice includes a small term in the denominator to avoid division by zero
    \item \textit{Gradient Behavior}: Backpropagation dynamics differ during training
\end{itemize}
\section{Experimental Results}
\label{sec: results}
\subsection{Quantitative Performance Evaluation}
A comprehensive performance evaluation was performed to rigorously assess the capabilities of the proposed SAGE-GAN model. From a total dataset comprising 140 image-mask pairs, a subset of 28 pairs, comprising 20\% of the total data, was randomly sampled and exclusively allocated for validation purposes. This validation set remained unutilized throughout all phases of model training, thereby ensuring an unbiased evaluation of the model's capacity for generalization. The remaining 112 image-mask pairs were used solely for the training of the model throughout its two distinct developmental stages.
\subsubsection*{Segmentation Results\vspace{2pt}\\}
The training process started with a 200-epoch pretraining phase using the Attention U-Net architecture \cite{oktay2018attentionunetlearninglook} to build an initial understanding of the feature space. This was followed by a 500-epoch training phase, embedding the pretrained Attention U-net into the   CycleGAN architecture \cite{zhu2020unpairedimagetoimagetranslationusing}, tailored to the S1 nanoparticle dataset, to enhance the model’s generative and discriminative capabilities for accurate segmentation.
\begin{figure*}[htbp]
    \centering    
    % First row of images
    \begin{minipage}[t]{0.3\textwidth}
    \textbf{}
        \centering\\
        \vspace{1mm}
        \includegraphics[width=\linewidth, height=3cm]{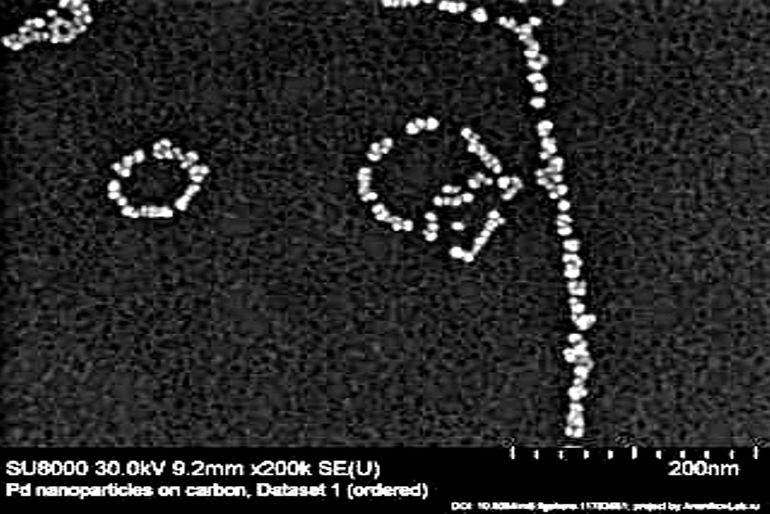}

        \vspace{1.4mm}
        
        \includegraphics[width=\linewidth, height=3cm]{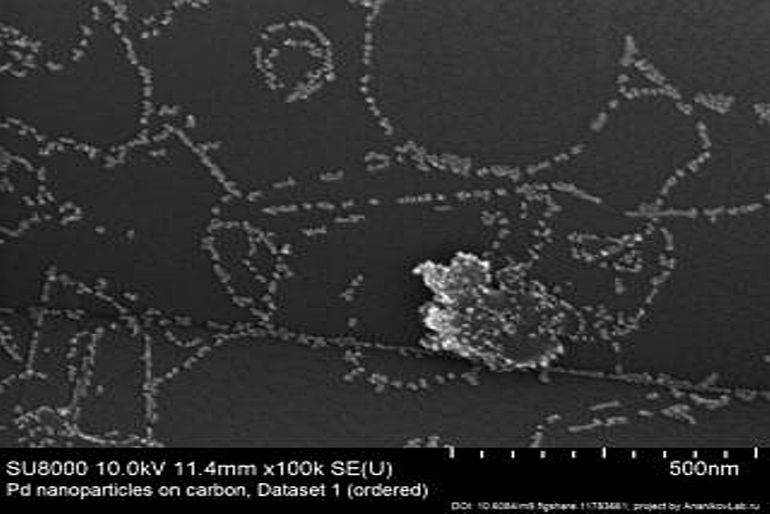}

        \vspace{1.4mm}
        
        \includegraphics[width=\linewidth, height=3cm]{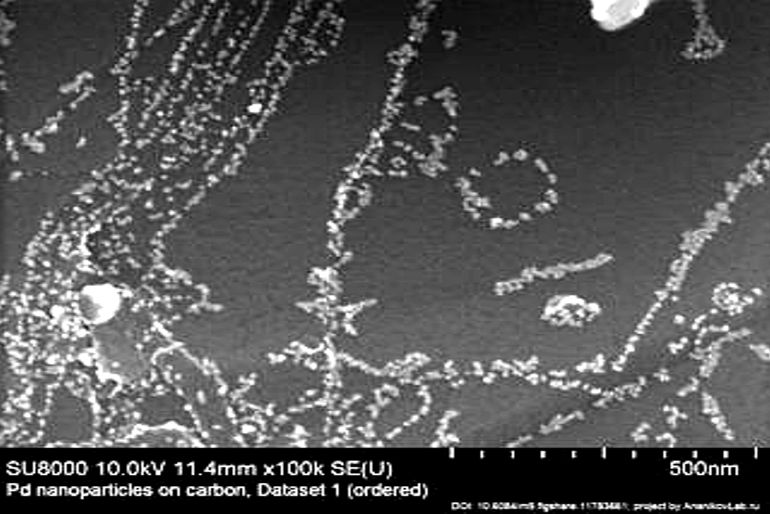}
        \subcaption{Real Image}
        \label{fig:typeA}
    \end{minipage}\hspace{0.005\textwidth}
    \begin{minipage}[t]{0.3\textwidth}
        \centering
        \textbf{}\\
        \vspace{1mm}
        \includegraphics[width=\linewidth, height=3cm]{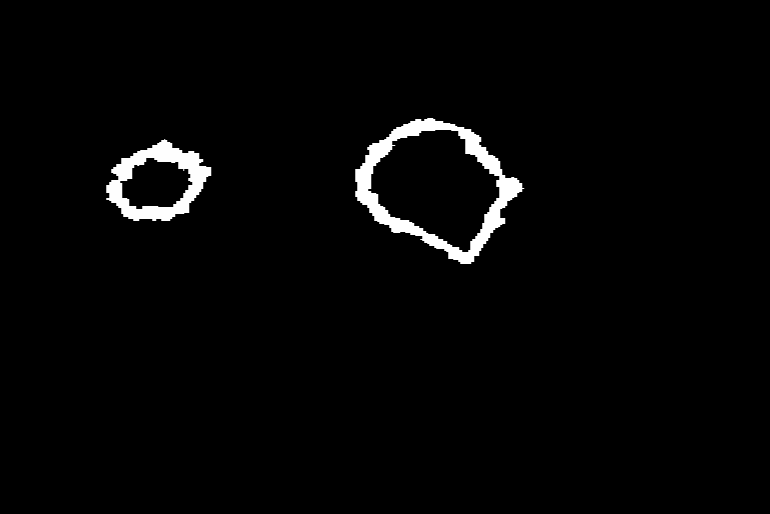}

        \vspace{1.4mm}
        
        \includegraphics[width=\linewidth, height=3cm]{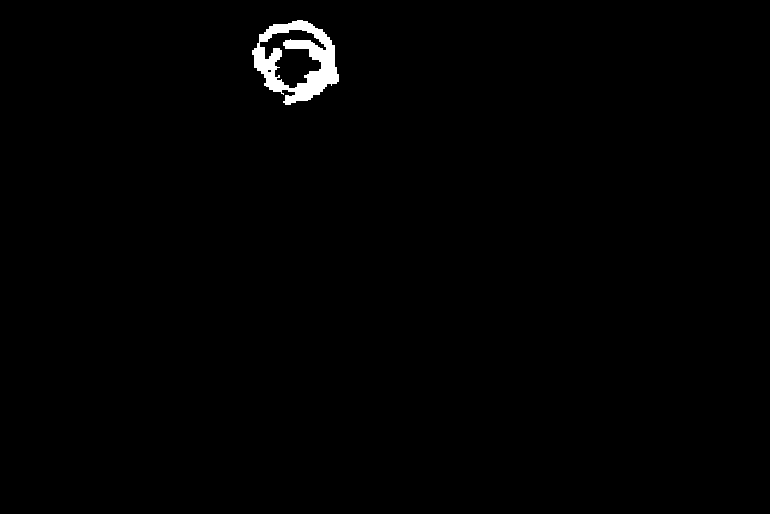}

        \vspace{1.4mm}
        
        \includegraphics[width=\linewidth, height=3cm]{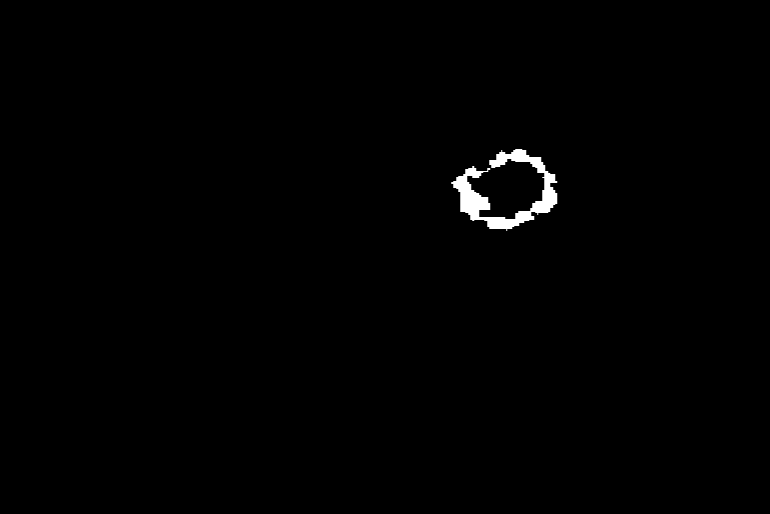}
        \subcaption{Real Mask}
        \label{fig:typeB}
    \end{minipage}\hspace{0.005\textwidth}
    \begin{minipage}[t]{0.3\textwidth}
        
        \vspace{1mm}
        
        \includegraphics[width=\linewidth, height=3cm]{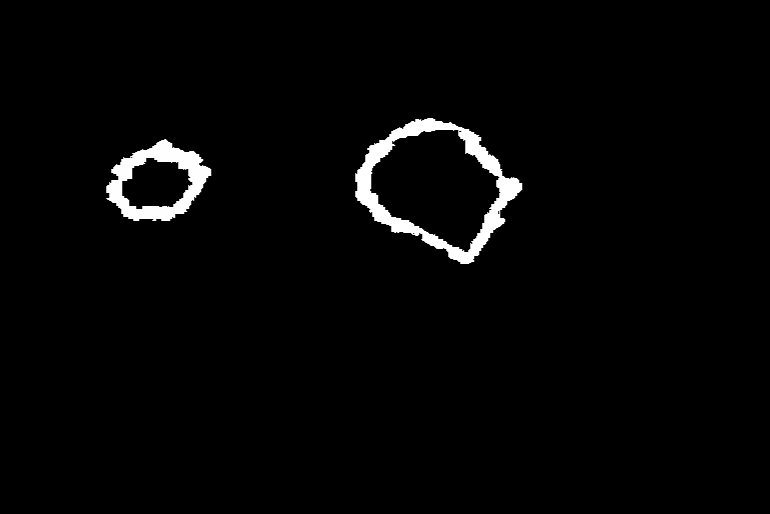}

        \vspace{1.4mm}
        
        \includegraphics[width=\linewidth, height=3cm]{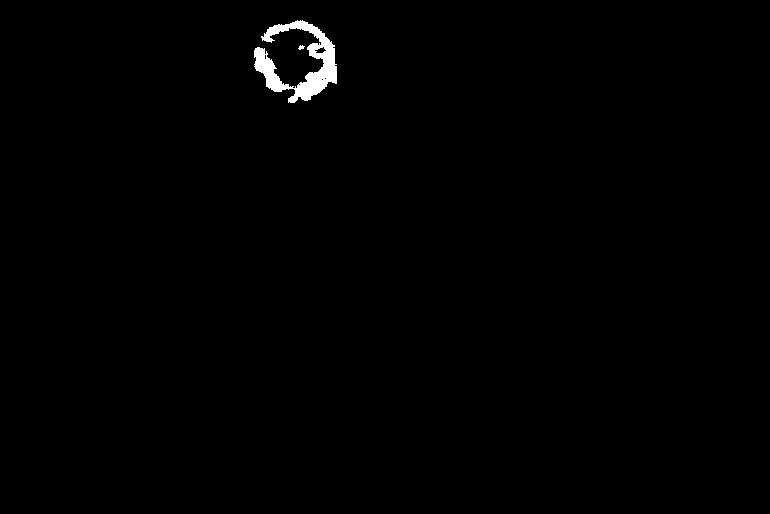}

        \vspace{1.4mm}
        
        \includegraphics[width=\linewidth, height=3cm]{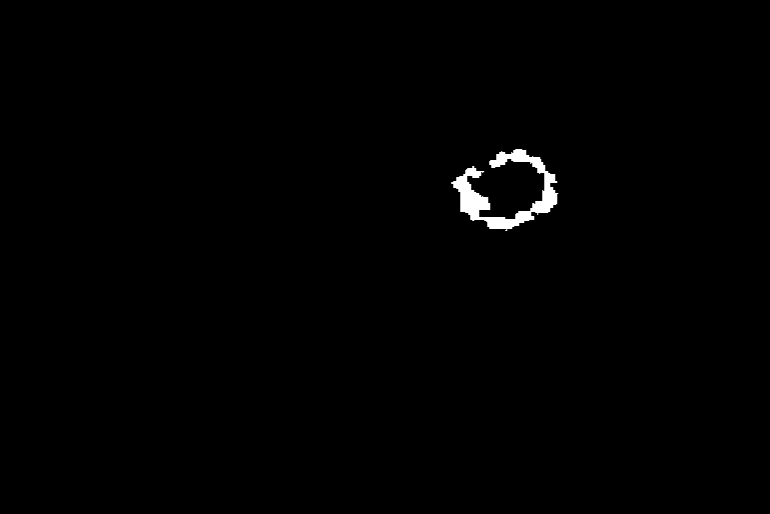}
        \subcaption{Predicted Mask}
        \label{fig:typeC}
    \end{minipage}\hspace{0.005\textwidth} 
    \caption{ Segmentation results using the proposed SAGE-GAN model. Each triplet displays (from left to right): the original SEM image, the manually annotated ground truth mask, and the corresponding predicted mask generated by SAGE-GAN. The model demonstrates strong agreement with the ground truth, effectively capturing nanoparticle boundaries and preserving structural detail.}
    \label{fig: seg_pred}
\end{figure*}
Upon completion of this extensive training process, the SAGE-GAN model exhibited exceptional segmentation performance when applied to the previously unseen validation set. This performance was quantified using the two chosen validation metrics, i.e., the Dice Similarity Coefficient (DSC) and the F1-score. The model achieved a notable Dice score of \textbf{0.932}, signifying a high degree of concordance between the predicted segmentation masks and the ground truth annotations. Furthermore, an F1-score of \textbf{0.956} was achieved, reflecting an excellent equilibrium between precision and recall in the identification of loop-like nanostructures highlighted in Fig-\ref{fig: seg_pred}.
\subsubsection*{Comparative Models\vspace{2pt}\\}
As shown in Table-\ref{tab:comparative models}, a comparative analysis was performed against several recent state-of-the-art segmentation architectures. The comparative data clearly highlights the superior efficacy of the SAGE-GAN model. The SAGE-GAN model significantly outperformed all baseline methods across both evaluation metrics. Notably, it achieved a 7.25\% improvement in Dice score and an 9.25\% gain in F1 score compared to its closest competitor, the Attention U-Net. This substantial performance gap highlights SAGE-GAN’s superior ability to accurately segment complex morphological images like S1. These results underscore the effectiveness of the model’s architectural enhancements and training strategy, likely driven by the powerful integration of self-attention mechanisms and the GAN framework, which together enhance feature extraction and contextual understanding for improved segmentation. 
\begin{table}[ht]
\centering
\caption{Comparative Models}
\vspace{1mm}

%\resizebox{\textwidth}{!}{ % Adjust width and height as needed
\begin{tabular}{|p{3cm}|p{2cm}|p{2cm}|}
\hline\centering
\textbf{Models} & \centering\textbf{Dice Score} & \centering\textbf{F1 Score} \tabularnewline\hline\centering
Half U-Net & \centering 0.595 & \centering 0.487 \tabularnewline\hline\centering
DC U-Net & \centering 0.584 & \centering 0.592 \tabularnewline\hline\centering
U-Net & \centering 0.681 & \centering 0.688 \tabularnewline\hline\centering
U-Net++ & \centering 0.698 & \centering 0.682 \tabularnewline\hline\centering
cGAN-Seg & \centering 0.865 & \centering 0.862 \tabularnewline\hline\centering
Attention U-Net & \centering 0.869 & \centering 0.875 \tabularnewline\hline\centering
\textbf{SAGE-GAN} & \centering\textbf{0.932} & \centering\textbf{0.956} \tabularnewline\hline
\end{tabular}
%}
\label{tab:comparative models}
\end{table}
\subsection*{Ablation Study}
Integral to the model's success was the custom choice of the loss function. A carefully formulated hybrid loss was employed, combining the strengths of Cross Entropy with a Focal Tversky loss\cite{abraham2018novelfocaltverskyloss}(parameterized with $\alpha=0.3, \beta=0.7, \gamma=1.5$). This composite loss function demonstrated markedly superior performance compared to the utilization of the standard Dice loss in isolation. Specifically, the hybrid loss yielded an improvement of 4.25\% in the Dice score and a substantial enhancement of 11.68\% in the F1-score. The advantages of this tailored loss function were particularly evident in its capacity for precise boundary capture and effective detection of smaller nanoparticle instances. The improved performance stems from the model's ability to effectively address class imbalance and accurately capture challenging edge pixels. By integrating these crucial features, the SAGE-GAN model delivers more precise and dependable segmentation results.
\begin{comment}
\begin{table}[h]
\centering
\caption{Ablation Study}
\vspace{1mm}

\begin{tabular}{|p{5cm}|p{2.5cm}|p{2.5cm}|}
\hline\centering
\textbf{Loss Function} & \centering\textbf{Dice Score} & \centering\textbf{F1 Score} \tabularnewline\hline\centering
Dice Loss & \centering 0.894 & \centering 0.856 \tabularnewline\hline\centering
CE + FTV & \centering \textbf{0.932} & \centering \textbf{0.956} \tabularnewline\hline
\end{tabular}
\label{tab:ablation}
\end{table}
\end{comment}
%
\subsection{Qualitative Evaluation: Attention Mechanism Analysis}
\begin{figure*}[h!]
    \centering    
    % First row of images
    \begin{minipage}[t]{0.3\textwidth}
    \textbf{}
        \centering\\
        \vspace{1mm}
        \includegraphics[width=\linewidth, height=3cm]{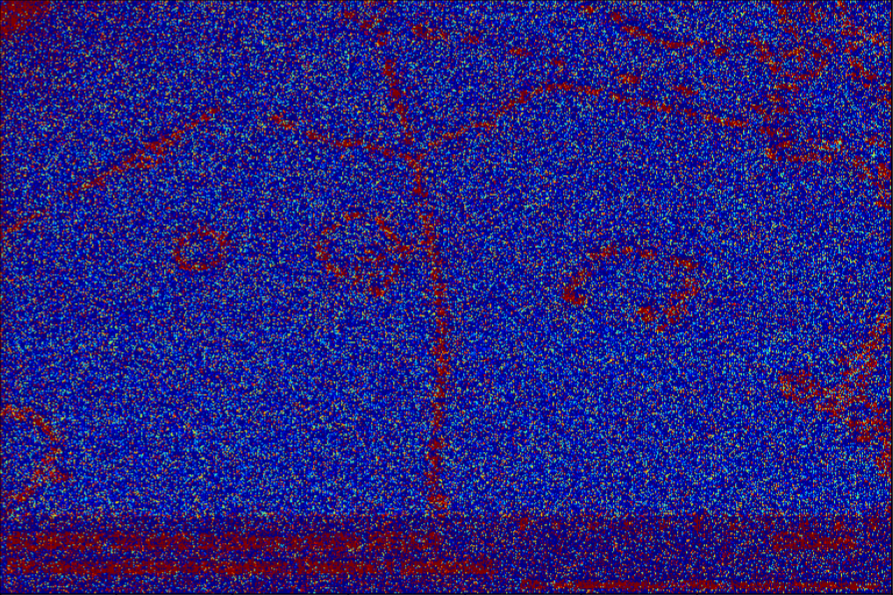}

        \vspace{1.4mm}
        
        \includegraphics[width=\linewidth, height=3cm]{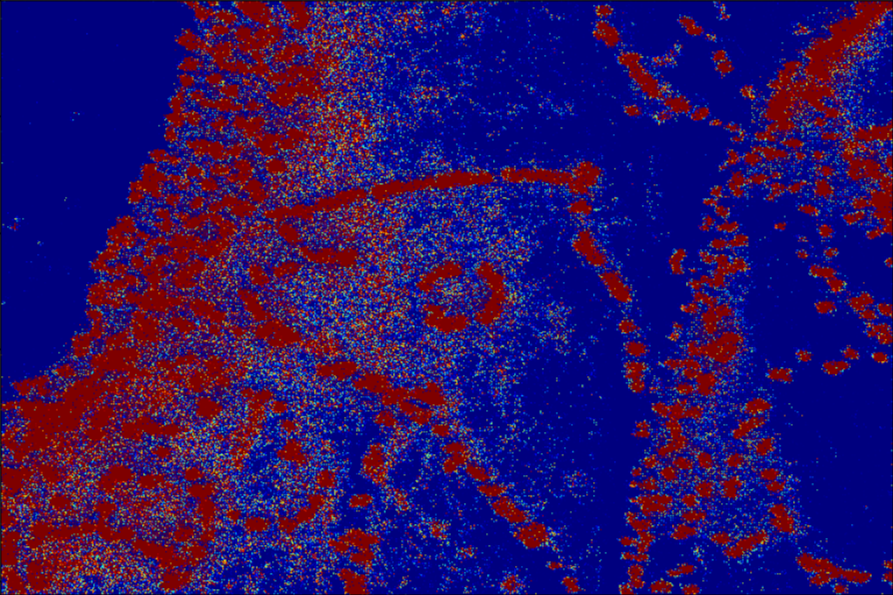}
        \subcaption{Initial Stage}
        \label{fig:attn_eval_initial}
    \end{minipage}\hspace{0.005\textwidth}
    \begin{minipage}[t]{0.3\textwidth}
        \centering
        \textbf{}\\
        \vspace{1mm}
        \includegraphics[width=\linewidth, height=3cm]{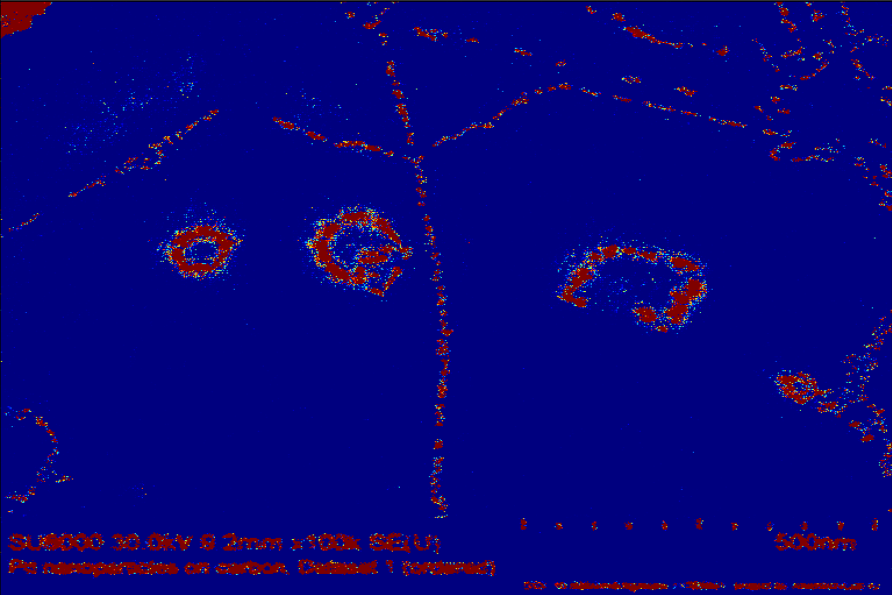}

        \vspace{1.4mm}
        
        \includegraphics[width=\linewidth, height=3cm]{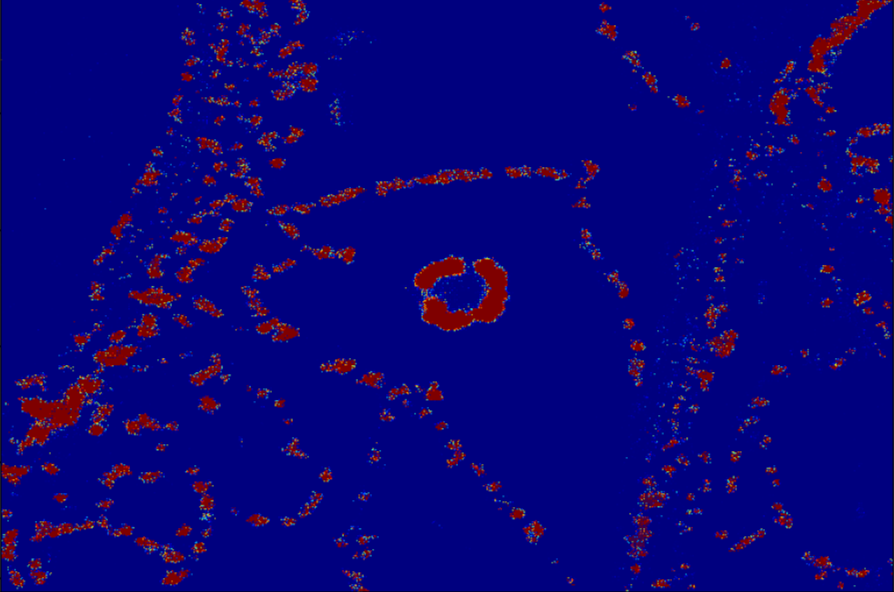}
        \subcaption{Intermediate Stage}
        \label{fig:attn_eval_intermediate}
    \end{minipage}\hspace{0.005\textwidth}
    \begin{minipage}[t]{0.3\textwidth}
        \centering
        \textbf{}\\
        \vspace{1mm}
        \includegraphics[width=\linewidth, height=3cm]{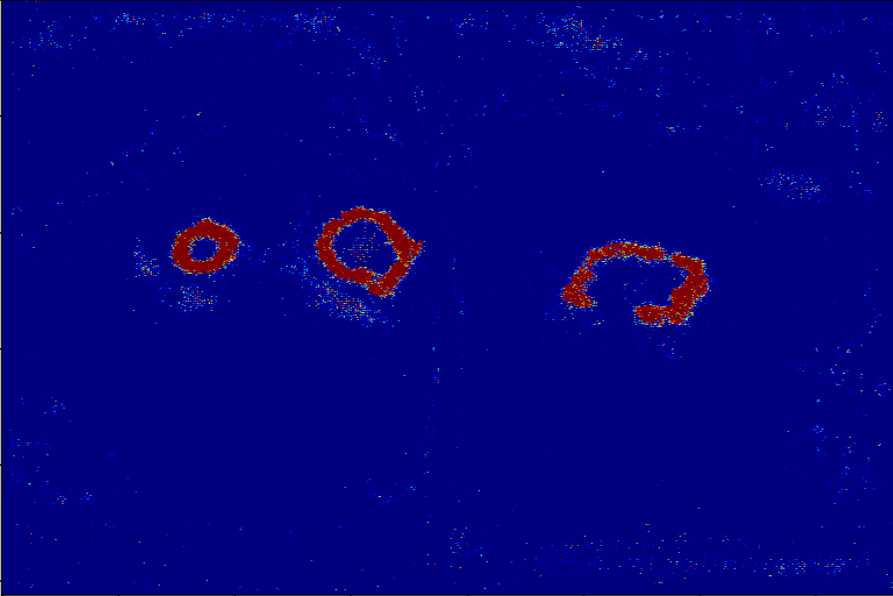}

        \vspace{1.4mm}
        
        \includegraphics[width=\linewidth, height=3cm]{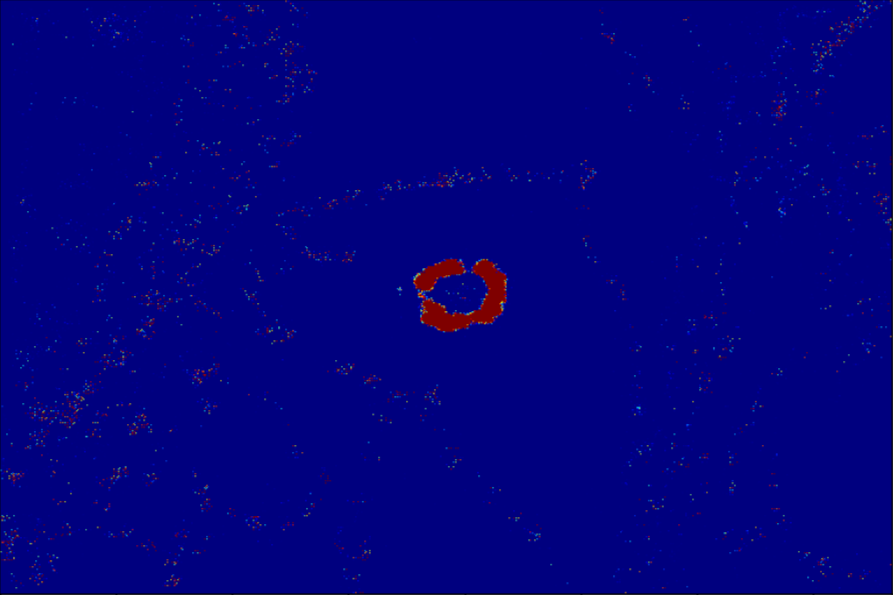}
        \subcaption{Final Stage}
        \label{fig:attn_eval_final}
    \end{minipage}\hspace{0.005\textwidth} 
    \caption{ Evolution of attention coefficients during training. The heatmaps visualize the output of Attention Gates (AG) at different training epochs, progressing from initial (left) to later epochs (right). Warmer regions (red) indicate areas of higher focus, highlighting how the model progressively learns to localize salient nanoparticle regions more effectively over time.}
    \label{fig:attn_evl}
\end{figure*}

%\subsubsection*{Attention Mechanism Analysis\vspace{2pt}\\}
%
In addition to evaluating the operational dynamics of our model, Fig-\ref{fig:attn_evl} illustrates the progressive evolution of the attention coefficients produced by the Attention Gates(AG) during the training process. The attention coefficients \cite{vaswani2017attention} are visualized here as a heat map, where red indicates higher focus compared to the colder regions depicted in blue.
%%%%
These visualizations provide critical insights into how the model learns to discern and prioritize salient regions in the image. Initially, as depicted in Figure-\ref{fig:attn_eval_initial}, the attention map exhibited a diffuse and relatively noisy activation pattern in the input image, indicative of the model's initial stage of feature learning. However, as training progressed through intermediate stages (Fig-\ref{fig:attn_eval_intermediate}), a visible refinement was observed in the attention mechanism, with activations beginning to concentrate on areas relevant to loop-like structures. In the final stages of training (Fig-\ref{fig:attn_eval_final}), the attention maps demonstrated a highly precise and focused localization of attention, predominantly at the boundaries of the target nanoparticles. This adaptive focusing capability is depicted as a key contributor to the enhanced performance of the model. The demonstrable ability of the model to dynamically adjust its attentional focus underpins its significant improvement in segmentation accuracy over standard U-Net \cite{ronneberger2015unetconvolutionalnetworksbiomedical} architectures when faced with such complex visual data.
\begin{figure}[htbp]
    \centering
    % Create a container that fits in one column
    \begin{minipage}{\linewidth}
        % First row: labels
        \begin{minipage}[t]{0.49\linewidth}
            \centering
            %\textbf{(a) Real Images}
        \end{minipage}
        \hfill
        \begin{minipage}[t]{0.49\linewidth}
            \centering
            %\textbf{(b) Attention Overlays}
        \end{minipage}
        
        \vspace{1mm}
        
        % Second row: first image pair
        \begin{minipage}[t]{0.49\linewidth}
            \centering
            \includegraphics[width=1\linewidth]{real_img_1.png}
        \end{minipage}
        \hfill
        \begin{minipage}[t]{0.49\linewidth}
            \centering
            \includegraphics[width=1\linewidth]{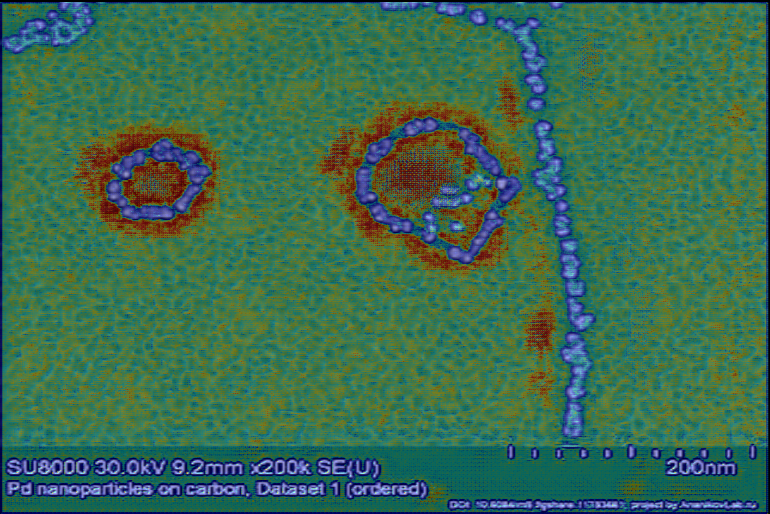}
        \end{minipage}
        
        \vspace{1mm}
        
        % Third row: second image pair
        \begin{minipage}[t]{0.49\linewidth}
            \centering
            \includegraphics[width=1\linewidth]{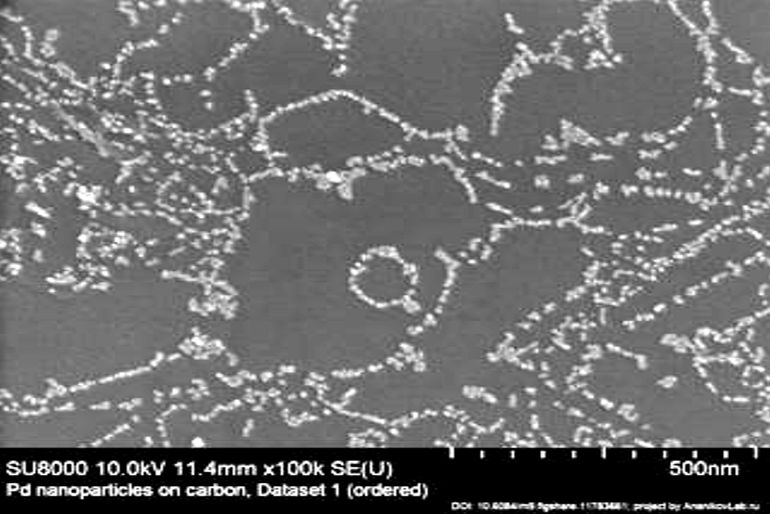}
            \subcaption{Real Images}
        \end{minipage}
        \hfill
        \begin{minipage}[t]{0.49\linewidth}
            \centering
            \includegraphics[width=1\linewidth]{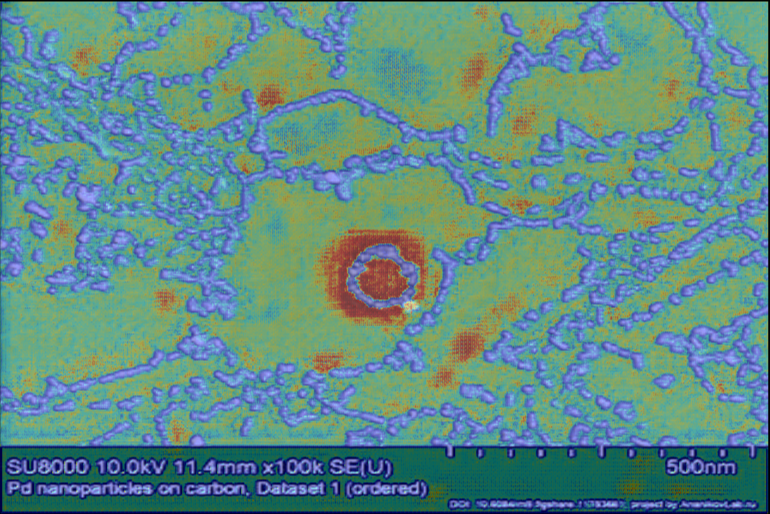}
            \subcaption{Attention Overlay}
        \end{minipage}
        
        \vspace{1mm}
        
        % Caption
        \caption{Attention overlay on validation set samples. Each pair displays the original SEM image with superimposed attention heatmaps generated by the Attention Gates. Darker regions indicate higher attention concentration, highlighting the model’s focus on nanoparticle boundaries and intricate morphological features. The overlays illustrate strong alignment with annotated masks, providing qualitative validation of the model’s segmentation accuracy and interpretability.}
        \label{fig:attn_overlay}
    \end{minipage}
\end{figure}
%%%%%%%%%%%%%%%%%%%%%%%%%%%%%%%%%%
The visualization of the attention overlay on the validation set, comparing real images and its counterpart attention overlays, demonstrates the model's precise focus on nanoparticle boundaries and complex morphological features. This is shown in Fig-\ref{fig:attn_overlay}. By superimposing attention heatmaps (showing high-concentration regions in darker colors) onto the original microscopy images, we observed strong alignment between predicted and ground truth masks, particularly in edge detection and irregular shapes. Minor discrepancies occur primarily in similar particle size and structure distributed across the whole image and on the boundaries, consistent with known segmentation challenges. These visualizations qualitatively illustrate the model's strong performance in terms of high Dice and F1 scores, offering intuitive insight into its decision-making process and confirming its capacity to generalize across complex morphologies without the need for specialized training modifications.
\section{Data and Code Availability}
\label{sec: code}
The code pertaining to this paper can be found at https://github.com/varunajith/SAGE-GAN. The dataset used in this study is publicly available at: \url{https://doi.org/10.1371/journal.pone.0311228.s001}.
\section{Training Configuration}
\label{sec: trainconfig}
All experiments were carried out on an Intel Xeon Silver 4216 CPU (2.10GHz), NVIDIA Quadro RTX 6000 GPU (24GB VRAM) and 64GB DDR4 RAM (2400MHz), with storage on Micron 1300 SATA SSDs (512GB).
\section{Conclusion}
\label{sec: conclusion}
Comprehension of nanoscale materials and analysis of corresponding microscope pictures necessitate substantial and time-consuming human involvement.  As a result, automatic identification and segmentation of nanoparticles in electron microscopy images has become an emergent area of research. When intricate patterns and intriguing ordering effects are included in these SEM images, the segmentation problem for both individual nanoparticles and arrays of nanoparticles becomes even more challenging. In this work, we introduced SAGE-GAN, a novel two-stage segmentation framework that integrates an Attention U-Net\cite{oktay2018attentionunetlearninglook} with a CycleGAN\cite{zhu2020unpairedimagetoimagetranslationusing}-based generative model to address the persistent challenges of nanoparticle segmentation in scanning electron microscopy (SEM) images.  The dataset we took into consideration is unique since it contains a few real SEM images in addition to offering the chance to investigate the unusual phenomena of nanoparticle ordering and hierarchical organization. The proposed architecture effectively mitigates the limitations of traditional deep learning models, particularly their reliance on large, annotated datasets, by generating structurally consistent synthetic images that augment the training process. The self-attention mechanism embedded within the U-Net\cite{ronneberger2015unetconvolutionalnetworksbiomedical} enables the model to focus selectively on morphologically relevant regions, enhancing segmentation accuracy even under noisy and low-contrast conditions. By coupling this with a segmentation-guided CycleGAN\cite{zhu2020unpairedimagetoimagetranslationusing}, the model not only learns the intricate features of nanoparticle structures, but also ensures that the generated synthetic data preserves both texture and geometric fidelity.\\
Experimental evaluations on the S1 nanoparticle dataset show that SAGE-GAN surpasses several state-of-the-art models, achieving a Dice score of 0.932 and an F1-score of 0.956. A comprehensive ablation study confirms the effectiveness of the hybrid loss function, while generalization tests on a separate subset of the graphene nanoparticle dataset demonstrate the model’s robustness in adapting to diverse nanoparticle morphologies (\textit{see} supplementary Fig-S2). Overall, SAGE-GAN provides a scalable, accurate, and generalizable solution for nanoparticle segmentation, making it a valuable tool for accelerating nanomaterial research, quality control, and high-throughput characterization pipelines in material science and nanotechnology.
\section*{Supplementary Information}
Supplementary figures, including synthetic generation results and generalization tests on graphene nanoparticle datasets, are provided in the supplementary material.
\section*{Author Contributions}
SG and SB conceptualized the idea. AP and VA were involved in methodology,software,validation and writing original draft. SB and SG provided supervision,formal analysis and review of the draft. All authors checked and edited the final manuscript.
%
\begin{comment}
Anindya Pal: Conceptualization, Methodology, Software, Validation, Writing -- Original Draft.\\
Varun Ajith: Conceptualization, Methodology, Visualization, Software, Writing -- Original Draft.\\
Saumik Bhattacharya: Supervision, Formal Analysis, Writing -- Review \& Editing.\\
Sayantari Ghosh: Supervision, Project Administration, Writing -- Review \& Editing.
\end{comment}
%
\section*{Conflicts of Interest or Competing Interests}
The authors declare no known conflict of interest.
\section*{Acknowledgments}
SG and SB  would like to acknowledge support from the ICTP (International Centre for Theoretical Physics) through the Associates Programme (2023–2028).
\section*{Declarations}
\subsection*{Ethical Approval}
Not applicable, as no living or human subjects were involved in the experiments.
\subsection*{Funding}
Not applicable.
{\small 
\bibliographystyle{splncs04}
\bibliography{bibliography}
}
\end{document}